\documentclass[10pt,twocolumn,letterpaper]{article}

\usepackage[algorithms]{wacv}      

\usepackage{graphicx}
\usepackage{booktabs}

\usepackage{multirow}
\usepackage{amssymb}
\usepackage{pifont}
\usepackage{xcolor}
\usepackage{algorithm}
\usepackage{algpseudocode}
\usepackage{wrapfig}
\usepackage{fontawesome5}

\definecolor{wacvblue}{rgb}{0.21,0.49,0.74}
\usepackage[breaklinks,colorlinks,allcolors=wacvblue]{hyperref}

\def\wacvPaperID{3103} 
\def\confName{WACV}
\def\confYear{2027}

\title{ RelateGS: Reliable Semantic and Relational Evidence for 3D Scene Understanding with Gaussian Splatting }

\author{
\begin{tabular}{c}
Seongjun Choi \qquad Sangyoun Lee$^{\dagger}$ \\[3pt]
Yonsei University \\[2pt]
School of Electrical and Electronic Engineering \\[2pt]
{\tt\small sjchoi.dp@yonsei.ac.kr} \\[9pt]
\href{https://drawingprocess.github.io/RelateGS/}
{\faGlobe\ Project Page}
\qquad
\href{https://github.com/DrawingProcess/RelateGS}
{\faGithub\ Code}
\end{tabular}
}

\begin{document}

\maketitle

\begingroup
\renewcommand{\thefootnote}{\fnsymbol{footnote}}
\footnotetext[2]{Corresponding author.}
\endgroup

\begin{abstract}
Language-aware 3D Gaussian Splatting (3DGS) has advanced language-driven 3D scene understanding by incorporating semantic and relational cues into reconstructed scenes.
However, relation-aware grounding depends not only on representing such information, but also on whether the underlying scene-side evidence is reliable and sufficient.
Multi-view semantic aggregation can incorporate weak observations, while object relations can remain under-observed along the available camera trajectory, leading to ambiguous evidence for target disambiguation.
We present RelateGS, a structured language-aware Gaussian representation that selectively refines semantic and relational evidence for 3D visual grounding.
RelateGS suppresses unreliable observations during semantic feature uplifting and consolidates multi-view object observations into a persistent 3D graph.
The graph further identifies under-observed object relations and guides targeted re-querying of informative existing frames to complement their
evidence.
Experiments on Sr3D+ and Nr3D demonstrate competitive 3D grounding performance, improving the previous best by 7.9 and 1.7 percentage points at Acc@0.25, respectively.
\end{abstract}    
\section{Introduction}
\label{sec:introduction}

\begin{figure}[!t]
\centering
\includegraphics[width=\linewidth]{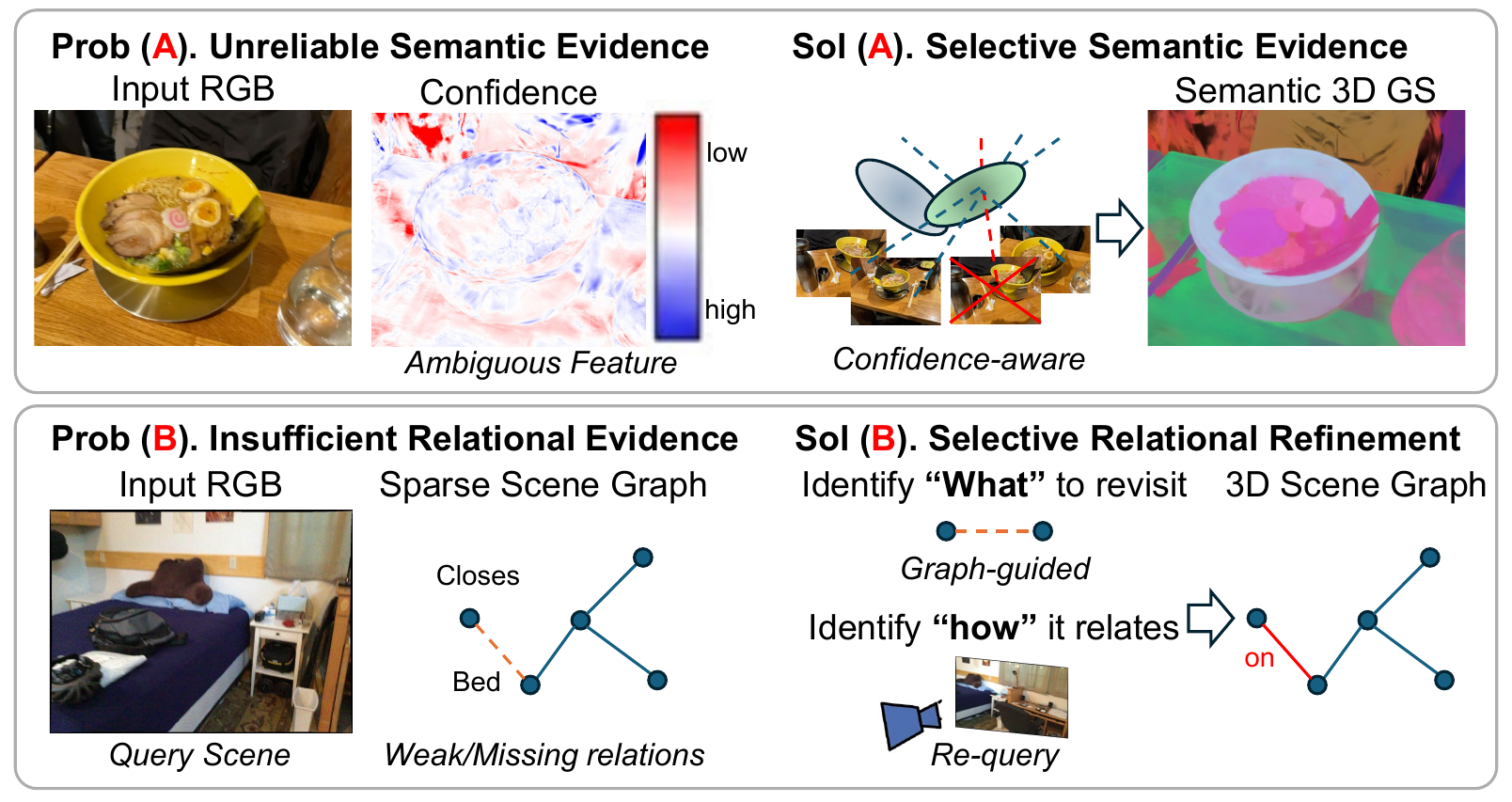}
\caption{ \textbf{Motivation of RelateGS.}
Existing methods can suffer from
(A) unreliable semantic evidence due to weak multi-view observations and
(B) insufficient relational evidence due to limited viewpoint coverage.
RelateGS addresses these issues by
(A) filtering weak observations during semantic feature uplifting and
(B) complementing under-observed relations through graph-guided re-querying
of informative existing views.
}
\label{fig:main_intro}
\end{figure}

Understanding and localizing objects in 3D environments from natural language
is essential for embodied applications such as navigation and manipulation.
3D Gaussian Splatting (3DGS)~\cite{3dgs} provides an explicit and
high-fidelity scene representation with real-time rendering, and recent
language-aware extensions associate Gaussian primitives with vision-language
features to support open-vocabulary localization and semantic
understanding~\cite{langsplat,LEGaussian,opengaussian,occamslgs}.
By embedding language-aligned semantics directly into an explicit 3D
representation, these approaches provide a promising foundation for grounding
natural-language descriptions in reconstructed 3D scenes.

Despite these advances, many referring expressions identify a target not only by \emph{what} it is, but also by \emph{how} it relates to surrounding objects.
While language-aware 3DGS can retrieve semantically relevant objects, multiple instances may appear equally plausible, yet only one may satisfy a description such as ``the chair next to the table.''
Resolving such expressions therefore requires object-level relational context beyond semantic similarity alone.

Recent approaches have extended language-aware 3D representations from dense semantics to structured relational representations by organizing reconstructed scenes into persistent object instances and modeling their pairwise relations~\cite{gaussiangraph, reasongrounder, chainsemantics}.
However, constructing an object-centric relational representation does not by itself guarantee reliable relational grounding, as shown in Fig.~\ref{fig:main_intro}.
The resulting representation is ultimately constrained by the quality and coverage of the visual observations from which its semantics and relations are constructed.
First, semantic evidence can be unreliable because language features are aggregated from observations affected by viewpoint changes, partial occlusion, object boundaries, and limited visibility. 
Weak observations can therefore corrupt the semantic representation of a Gaussian and increase ambiguity among similar object candidates.
Second, relational evidence can be insufficient because pairwise relations are inferred only from the object configurations exposed by the available camera views.
When a grounding-relevant object pair is poorly observed, the relation needed to distinguish the target may remain weak or missing.
Consequently, even an explicitly structured scene can fail to resolve relational language when its semantic evidence is unreliable or its relational evidence is under-observed.

Our key insight is that relation-aware grounding benefits from selectively refining the scene-side evidence supporting the representation, rather than treating all available observations as equally informative.
Specifically, weak semantic observations should be suppressed during Gaussian feature construction, while under-observed object relations should be complemented using informative existing observations.
Based on this insight, we introduce RelateGS, a structured language-aware Gaussian representation that selectively refines the scene-side evidence required for relation-aware 3D grounding.

Specifically, RelateGS realizes this principle through complementary semantic and relational
refinement.
First, \emph{confidence-aware semantic uplifting} estimates the contribution of each image observation to individual Gaussians and selectively aggregates reliable language evidence, reducing semantic ambiguity introduced by weak observations.
The resulting multi-view semantics are consolidated into a persistent 3D object graph that maintains object identities and spatial relationships across views.
Second, \emph{graph-guided relational re-querying} reasons over the current graph to identify under-observed object pairs and selects informative frames from the existing camera trajectory for targeted re-querying.
The recovered relational evidence is fused back into the persistent graph, allowing RelateGS to complement missing relations rather than treating the initially constructed graph as a fixed output.

Extensive experiments demonstrate that RelateGS substantially improves relation-aware 3D visual grounding.
On Sr3D+ and Nr3D~\cite{referit3d}, RelateGS achieves competitive overall grounding performance, showing the effectiveness of selectively refining scene-side semantic and relational evidence.
Additional semantic evaluations and controlled ablations further validate the
underlying representation and the contribution of each refinement strategy.

Our contributions are summarized as follows:
\begin{itemize}
    \item We formulate relation-aware 3D grounding from the perspective of
    scene-side evidence quality, considering both the reliability of
    Gaussian-level semantic evidence and the sufficiency of object-level
    relational evidence.

    \item We introduce RelateGS, which realizes selective evidence refinement
    through confidence-aware semantic uplifting and graph-guided relational
    re-querying over a persistent 3D object graph.

    \item RelateGS improves grounding performance on Sr3D+ and Nr3D,
    while complementary semantic evaluations and controlled ablations validate
    the proposed representation design.
\end{itemize}
\section{Related Work}
\label{sec:related_work}

\subsection{Language-Embedded 3D Gaussian Splatting}

Recent advances in 3DGS~\cite{3dgs} have extended
its role from photorealistic scene reconstruction to language-aware 3D
understanding. Building on 2D foundation models such as SAM~\cite{sam} and CLIP~\cite{clip}, recent approaches extract object masks and language-aligned features from multi-view images and transfer this semantic evidence to Gaussian primitives. LangSplat~\cite{langsplat} and
LEGaussians~\cite{LEGaussian} embed language-aligned features into Gaussian
scenes for open-vocabulary localization.
OpenGaussian~\cite{opengaussian} further targets point-level open-vocabulary
understanding through improved semantic representation and 2D--3D association.
GAGS~\cite{gags} improves multi-view semantic consistency through granularity-aware segmentation and feature distillation.
Instead of optimizing additional semantic parameters, LUDVIG~\cite{ludvig} and Occam's LGS~\cite{occamslgs} transfer multi-view language features to
pre-trained Gaussian primitives according to their rendering contributions.
More recently, ReasonGrounder~\cite{reasongrounder} introduces hierarchical
Gaussian feature fields together with large vision-language models for
open-vocabulary grounding and reasoning.

These methods demonstrate the effectiveness of dense language-aligned
Gaussian representations for connecting natural language with reconstructed
3D scenes.
However, the resulting Gaussian-level semantics ultimately depend on the
multi-view observations from which they are constructed.
Observations can vary substantially in reliability due to visibility,
occlusion, and object boundaries, yet this variation is not always explicitly
considered during semantic aggregation.
RelateGS builds on language-embedded Gaussians from this perspective,
focusing on how reliable semantic evidence can be selectively preserved for
downstream relation-aware grounding.

\subsection{3D Scene Graph Generation}

3D scene graphs provide an object-centric abstraction of a scene by
representing objects as nodes and their semantic or spatial relationships as
edges.
3DSSG~\cite{3dssg} establishes object and relation prediction from 3D point
clouds, while SceneGraphFusion~\cite{scenegraphfusion} incrementally constructs
scene graphs from RGB-D sequences.
Subsequent works improve the quality and consistency of object-level
representations: VL-SAT~\cite{vlsat} incorporates visual-linguistic context for
3D scene graph prediction, and SGAligner~\cite{sgaligner} exploits cross-modal
alignment for object association across 3D scenes.
With the emergence of 3DGS, GaussianGraph~\cite{gaussiangraph} extends this
object-centric perspective to Gaussian scenes by organizing reconstructed
primitives into objects with semantic attributes and spatial relationships.

These works demonstrate that object-centric graphs complement dense
semantic representations by explicitly modeling inter-object relationships.
However, the presence of a scene graph does not guarantee sufficient
observational support for relation-aware grounding.
In particular, limited visibility and co-visibility along the camera
trajectory can leave relational evidence incomplete, even when the
corresponding objects are represented in the graph.
RelateGS therefore leverages a persistent scene graph not only to
represent object relationships but also to identify under-observed
object pairs and guide targeted retrieval of complementary evidence
from multi-view observations.

\subsection{3D Visual Grounding}

3D visual grounding aims to localize a target object from a natural-language
referring expression.
When multiple instances share the same semantic category, grounding often requires spatial and contextual relationships with surrounding
objects.
Point-cloud-based approaches such as CORE-3DVG~\cite{core3dvg} and
ViewSRD~\cite{viewsrd} exploit contextual and relational cues to resolve such
ambiguity. ConcreteNet~\cite{concretenet} improves dense 3D grounding through relation-aware attention, contrastive learning, viewpoint modeling, and multi-view ensembling.
More recently, EG-3DVG~\cite{eg3dvg} introduces an expression- and
geometry-aware grounding decoder that improves text--vision alignment and
selectively aggregates geometry-consistent visual cues.

Recent methods exploit relational scene representations to disambiguate
referred objects~\cite{reasongrounder, gaussiangraph}.
Chain of Semantics Programming~\cite{chainsemantics} formulates grounding
as zero-shot neuro-symbolic reasoning, constructing a relationship graph
and chain of semantics while leveraging multi-view 3DGS rendering
for complementary semantic observations.
In contrast, RelateGS focuses on whether the scene representation
provides reliable and sufficient grounding evidence, rather than
introducing new decoders, reasoning programs, or rendered viewpoints.
It selectively preserves reliable Gaussian-level semantics and
complements under-observed object-level relations before downstream reasoning.
\section{Method}
\label{sec:method}

\begin{figure*}[t]
    \centering
    \includegraphics[width=\linewidth]{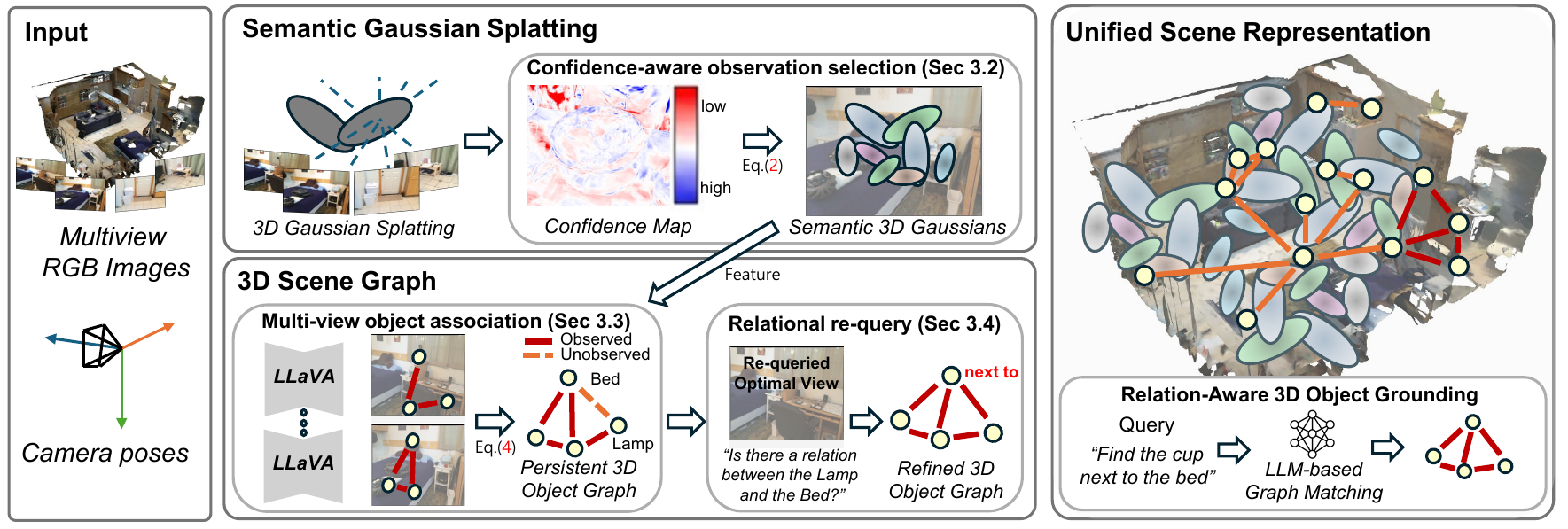}
    \caption{
    Overview of RelateGS.
    Given a reconstructed 3D Gaussian scene and posed RGB-D observations,
    RelateGS selectively refines the scene-side evidence used for
    relation-aware grounding.
    At the Gaussian level, weak multi-view observations are filtered during
    semantic uplifting.
    At the object level, multi-view observations are consolidated into a
    persistent graph, whose current state identifies under-observed object
    pairs and guides targeted re-querying of informative existing frames.
    }
    \label{fig:relategs_overview}
\end{figure*}

\subsection{Overview}
\label{sec:overview}

We present RelateGS, a structured language-aware Gaussian representation for
relation-aware 3D grounding.
Given posed RGB-D observations and a reconstructed 3D Gaussian scene, we construct semantic and relational evidence to disambiguate the referred object through its relations to other instances.

We first construct reliable Gaussian-level semantic evidence from multi-view observations and associate object observations across views to form an initial object graph. We then identify under-observed object pairs and selectively re-query informative observations to complement missing relational evidence, yielding a refined object graph for language grounding.

\subsection{Reliable Gaussian-Level Semantic Evidence}
\label{sec:semantic_evidence}

We first construct language-aligned Gaussian features from multi-view image
observations.
Our formulation builds on the training-free uplifting strategy of
Occam's LGS~\cite{occamslgs}, while explicitly accounting for the reliability
of the observations contributing to each Gaussian.



\noindent\textbf{Semantic feature uplifting.} Let $\mathbf{f}_{iv}$ denote the language feature associated with Gaussian $g_i$ from view $v$, and let $w_{iv}$ denote its rendering contribution. Following rendering-based semantic uplifting, the feature of $g_i$ is aggregated across its supporting views as: \begin{equation} \mathbf{f}_i = \frac{ \sum_{v \in \mathcal{V}_i} w_{iv}\mathbf{f}_{iv} }{ \sum_{v \in \mathcal{V}_i} w_{iv} }, \label{eq:semantic_uplifting} \end{equation} where $\mathcal{V}_i$ denotes the observations supporting $g_i$. This associates 2D language evidence with the reconstructed Gaussian primitives without additional semantic optimization. 

\noindent\textbf{Confidence-aware observation selection.}
Not every observation contributing to Eq.~\ref{eq:semantic_uplifting}
provides equally reliable evidence.
Partial occlusion or limited visibility can weaken a Gaussian's contribution
to a view, introducing ambiguous semantic evidence during multi-view
aggregation.
For each Gaussian $g_i$ observed from view $v$, we obtain its rendering
significance $s_{iv}$, derived during alpha compositing from its opacity
contribution and accumulated transmittance.
We use $s_{iv}$ as an observation-level confidence signal and retain only
sufficiently reliable observations:
\begin{equation}
    \tilde{w}_{iv}
    =
    w_{iv}\,
    \mathbb{I}\left(s_{iv} \geq \tau_c\right),
    \label{eq:confidence_weight}
\end{equation}
where $\tau_c$ denotes the confidence threshold.
The final language feature is obtained by replacing $w_{iv}$ in
Eq.~\ref{eq:semantic_uplifting} with $\tilde{w}_{iv}$, thereby suppressing
weak or occluded semantic observations.

\subsection{Persistent Object-Level Relational Evidence}
\label{sec:object_evidence}

Dense Gaussian features provide local semantic evidence, whereas relational
grounding requires persistent object identities and evidence describing how
those objects relate to one another.
We therefore consolidate frame-level observations into a persistent object
graph
\begin{equation}
    \mathcal{S}^{\mathrm{obs}}
    =
    (\mathcal{V},\mathcal{E}^{\mathrm{obs}}),
\end{equation}
where $\mathcal{V}$ contains persistent 3D object instances and
$\mathcal{E}^{\mathrm{obs}}$ contains observed relational edges.
These edges are constructed by associating VLM-predicted
relations with persistent object identities, complementing them with
geometry-derived relations, and filtering unreliable edges.

\noindent\textbf{Frame-level observations and 3D uplifting.}
For each RGB frame, we obtain view-level object vertices and pairwise relation
edges using LLaVA-SpaceSGG~\cite{llavaspacesgg}.
We refer to its outputs simply as VLM observations hereafter.

Each detected object is lifted into the reconstructed scene using the
corresponding depth map and calibrated camera pose.
Back-projected 3D points provide the spatial support of the observation and
are used to estimate its 3D center and extent and to associate it with the
Gaussian scene.
This produces a sequence of view-dependent 3D object observations along the
camera trajectory.

\noindent\textbf{Multi-view object association.}
The same physical object can appear repeatedly across frames.
To maintain consistent object identities, each incoming 3D object observation
$o_j$ is compared with existing persistent nodes $v_i$ using spatial,
semantic, and scale consistency.
We define the association score as
\begin{equation}
A_{ij}
=
\lambda_s A^{\mathrm{spatial}}_{ij}
+
\lambda_m A^{\mathrm{semantic}}_{ij}
+
\lambda_l A^{\mathrm{scale}}_{ij},
\label{eq:association}
\end{equation}
where $\lambda_s$, $\lambda_m$, and $\lambda_l$ weight the spatial,
semantic, and scale consistency terms, respectively.
The individual consistency terms are defined as
\begin{gather}
A^{\mathrm{spatial}}_{ij}
=
\exp\left(
-\frac{
\|\boldsymbol{\mu}_i-\boldsymbol{\mu}_j\|_2
}{
\sigma_s
}
\right),
\nonumber\\
A^{\mathrm{semantic}}_{ij}
=
\cos(\mathbf{e}_i,\mathbf{e}_j),
\nonumber\\
A^{\mathrm{scale}}_{ij}
=
\exp\left(
-\frac{
\left|
\|\mathbf{l}_i\|_2-\|\mathbf{l}_j\|_2
\right|
}{
\|\mathbf{l}_j\|_2+\epsilon
}
\right).
\label{eq:association_components}
\end{gather}
where $\boldsymbol{\mu}_i$ and $\mathbf{l}_i$ denote the 3D center and spatial extent of object $i$, respectively, and $\sigma_s$ is the spatial distance scale. The extent $\mathbf{l}_i$ contains the side lengths of the axis-aligned 3D bounding box enclosing the Gaussians associated with the back-projected 2D detection. $\mathbf{e}_i$ denotes the object-level semantic embedding aggregated from the language-aligned Gaussian features constructed in Sec.~\ref{sec:semantic_evidence}.

We apply Hungarian assignment to the resulting pairwise scores to obtain one-to-one correspondences between incoming observations and persistent nodes.
An assigned observation is merged into the corresponding node when
$A_{ij}\geq\tau_{\mathrm{assoc}}$; otherwise, it initializes a new persistent node.
Through this process, independently observed frame-level detections are consolidated into persistent object identities shared across views.

\noindent\textbf{Relational evidence fusion.}
Frame-level relations are transferred to the persistent graph through the object associations established above. Specifically, a relation predicted between two frame-local objects is remapped to their corresponding persistent node IDs and accumulated as a triplet
$(v_i,r,v_j)$ across views.
Thus, repeated observations of the same physical object pair contribute relational evidence to the same persistent endpoints.

After all frame-level observations are associated, we complement the
accumulated VLM relations with geometric relations inferred from the final
persistent node geometry.
Relations sharing the same subject, predicate, and object are fused across
views using the relation confidence provided by the VLM and their multi-view
support.
Weak or geometrically inconsistent edges are subsequently filtered to obtain
$\mathcal{E}^{\mathrm{obs}}$.
The resulting graph therefore preserves multi-view relational evidence while
maintaining consistency with the consolidated 3D object representation.
However, relational evidence can remain incomplete for object pairs that are
insufficiently observed across the input views.

\subsection{Graph-Guided Relational Refinement}
\label{sec:relation_refinement}

To complement such insufficient relational evidence, RelateGS uses the current
object graph to determine \emph{which object pairs require additional
relational evidence} and scene geometry to determine \emph{which existing
frames are informative for those pairs}.
The refinement consists of under-observed pair identification, informative
frame retrieval, and targeted relational re-querying.

\noindent\textbf{Identifying under-observed object pairs.}
We consider spatially nearby objects because their pairwise relations are
likely to provide useful cues for relational grounding.
Among these pairs, we identify those for which the accumulated VLM relation
is missing or remains below a confidence threshold.
Formally,
\begin{equation}
    \mathcal{P}_{\mathrm{under}}
    =
    \left\{
        (v_i,v_j)
        \;\middle|\;
        d(v_i,v_j) < \tau_d,\;
        c_{ij} < \tau_r
    \right\},
    \label{eq:under_observed}
\end{equation}
where $d(v_i,v_j)$ denotes the distance between the two persistent objects and $c_{ij}$ denotes the accumulated relation confidence obtained by fusing VLM-provided confidence across views.
Pairs without observed relational evidence are assigned zero confidence.
This criterion focuses subsequent refinement on nearby object pairs for which
the current graph provides insufficient relational evidence.

\noindent\textbf{Retrieving informative existing frames.}
For each pair $(v_i,v_j)\in\mathcal{P}_{\mathrm{under}}$, we search the original camera trajectory for a frame that provides a favorable joint view of both objects. We first project their 3D extents into each camera and retain only frames with sufficient joint frustum coverage. Among these candidates, we rank frames based on the camera-to-pair distance and the alignment of the viewing direction with the center of the object pair.
The highest-ranked frame is selected for targeted relational re-querying.

\noindent\textbf{Targeted relational re-query.}
The selected frame is re-queried with a pair-specific prompt asking for the
spatial relationship between the two under-observed objects.
Unlike the generic frame-level extraction used during initial graph
construction, this query explicitly targets a relational deficiency identified
in the current object graph.

The resulting observations are processed through the same 3D uplifting,
persistent association, and relation-fusion procedure used during initial
graph construction.
This ensures that the newly acquired relational evidence is associated with
the corresponding persistent object pair rather than treated as an independent
prediction.
After refinement, we obtain the refined object graph
\begin{equation}
\mathcal{S}^{\mathrm{ref}}
=
(\mathcal{V},\mathcal{E}^{\mathrm{ref}}),
\end{equation}
where $\mathcal{E}^{\mathrm{ref}}$ contains the original relational evidence
together with the selectively acquired re-query observations.

\subsection{Relation-Aware 3D Grounding}
\label{sec:grounding}

The final RelateGS representation combines language-aligned Gaussian features
with persistent object identities and their refined relational evidence.
Given a referring expression, this representation provides both semantic cues for identifying plausible objects and relational cues for disambiguating among competing instances.

For downstream grounding, we adopt the LLM-based graph matching protocol of
GaussianGraph~\cite{gaussiangraph}.
The matcher compares the referring expression with the semantic attributes
and relations stored in the object graph and selects the object node that
best satisfies the description.
Using the same grounding procedure across all variants isolates the effect of
our scene-side evidence refinement: improvements arise from the representation
provided to the matcher rather than from introducing a new task-specific
grounding module.
\section{Experiments}
\label{sec:experiments}

We evaluate RelateGS primarily on relation-aware 3D visual grounding, where
both semantic object evidence and inter-object relations are required to
resolve natural-language referring expressions.
We further evaluate the underlying semantic representation and conduct
controlled ablations to examine how each form of selective evidence
refinement contributes to grounding performance.

\subsection{Experimental Setup}
\label{sec:experimental_setup}

\noindent\textbf{Datasets and tasks.}
Our primary evaluation is 3D visual grounding on Sr3D+ and Nr3D from
ReferIt3D~\cite{referit3d}, which are built on ScanNet~\cite{scannet}.
These benchmarks contain referring expressions that identify target objects
through semantic, spatial, and contextual descriptions.
Following the evaluation setting adopted by BBQ~\cite{bbq} and
GaussianGraph~\cite{gaussiangraph}, we use the same eight ScanNet indoor
scenes to enable direct comparison with prior structured 3D scene
representations.

To independently assess the Gaussian-level semantic evidence underlying our
object graph, we additionally evaluate semantic segmentation on Replica
~\cite{replica} and ScanNet, and open-vocabulary localization on the four
LERF-OVS scenes: \textit{ramen}, \textit{teatime},
\textit{waldo\_kitchen}, and \textit{figurines}.

\noindent\textbf{Evaluation metrics.}
For 3D visual grounding, we report Acc@0.1 and Acc@0.25 according to the 3D
overlap between the predicted and ground-truth target regions.
We additionally report the Easy and Hard subsets of Sr3D+ and Nr3D to examine
performance under different levels of object ambiguity.
For semantic segmentation, we report mean Intersection-over-Union (mIoU) and
mean class accuracy (mAcc).
For LERF-OVS, we follow its evaluation protocol and report Acc@0.25,
Acc@0.5, and mIoU.

\noindent\textbf{Implementation details.}
All experiments are conducted on a single NVIDIA A5000 GPU with 24\,GB
memory.
We construct language-aligned Gaussian features following the uplifting strategy of Occam's LGS~\cite{occamslgs}, together with our confidence-aware observation selection described in Sec.~\ref{sec:semantic_evidence}.
Frame-level object and relation observations are obtained using
LLaVA-SpaceSGG~\cite{llavaspacesgg} and consolidated into the persistent
object graph.
For relational refinement, under-observed object pairs are identified from
the initial graph and re-queried from informative frames selected from the
original camera trajectory.
All compared re-query variants use the same additional VLM query budget.
Further hyperparameters, prompts, and implementation details are provided in
the supplementary material.

\subsection{3D Visual Grounding}
\label{sec:grounding_results}

We first evaluate whether the refined scene-side evidence provided by RelateGS
improves the primary task of relation-aware 3D grounding.
Tables~\ref{tab:results_sr3d} and~\ref{tab:results_nr3d} compare RelateGS
with point-cloud- and Gaussian-based scene representations on Sr3D+ and Nr3D.

\begin{table}[!t]
\vspace{-0.8em}
\centering
\caption{
Comparison on the \textbf{Sr3D+} benchmark.
We report Acc@0.1 and Acc@0.25 (\%) for the Easy, Hard, and overall splits.
}
\scriptsize
\setlength{\tabcolsep}{2pt}
\renewcommand{\arraystretch}{0.97}
\begin{tabular}{lcccccc}
\toprule
\multirow{2}{*}{Method}
& \multicolumn{2}{c}{Easy}
& \multicolumn{2}{c}{Hard}
& \multicolumn{2}{c}{Overall} \\
\cmidrule(lr){2-3}
\cmidrule(lr){4-5}
\cmidrule(lr){6-7}
& A@0.1$\uparrow$ & A@0.25$\uparrow$
& A@0.1$\uparrow$ & A@0.25$\uparrow$
& A@0.1$\uparrow$ & A@0.25$\uparrow$ \\
\midrule
LangSplat~\cite{langsplat}
& 4.7 & 1.5 & 2.1 & 1.1 & 4.5 & 2.3 \\
OpenGaussian~\cite{opengaussian}
& 7.3 & 5.1 & 3.6 & 1.9 & 7.0 & 4.9 \\
ConceptGraph~\cite{conceptgraphs}
& 13.0 & 6.8 & 16.0 & 1.3 & 13.3 & 6.2 \\
OpenFusion~\cite{openfusion}
& 14.0 & 2.4 & 1.3 & 1.3 & 12.6 & 2.4 \\
GaussianGraph~\cite{gaussiangraph}
& 19.1 & 7.7 & \textbf{16.3} & 5.6 & 18.2 & 7.4 \\
\midrule
\textbf{RelateGS (Ours)}
& \textbf{22.2} & \textbf{15.2}
& 15.2 & \textbf{6.6}
& \textbf{22.2} & \textbf{15.3} \\

\bottomrule
\end{tabular}
\label{tab:results_sr3d}
\end{table}

\begin{table}[!t]
\vspace{-0.8em}
\centering
\caption{
Comparison on the \textbf{Nr3D} benchmark.
We report Acc@0.1 and Acc@0.25 (\%) for the Easy, Hard, and overall splits.
}
\scriptsize
\setlength{\tabcolsep}{2pt}
\renewcommand{\arraystretch}{0.97}
\begin{tabular}{lcccccc}
\toprule
\multirow{2}{*}{Method}
& \multicolumn{2}{c}{Easy}
& \multicolumn{2}{c}{Hard}
& \multicolumn{2}{c}{Overall} \\
\cmidrule(lr){2-3}
\cmidrule(lr){4-5}
\cmidrule(lr){6-7}
& A@0.1$\uparrow$ & A@0.25$\uparrow$
& A@0.1$\uparrow$ & A@0.25$\uparrow$
& A@0.1$\uparrow$ & A@0.25$\uparrow$ \\
\midrule
LangSplat~\cite{langsplat}
& 8.5 & 1.7 & 3.4 & 1.2 & 7.4 & 1.5 \\
OpenGaussian~\cite{opengaussian}
& 10.1 & 5.9 & 7.4 & 3.1 & 9.5 & 4.7 \\
ConceptGraph~\cite{conceptgraphs}
& 18.7 & 9.2 & 9.1 & 2.0 & 16.0 & 7.2 \\
OpenFusion~\cite{openfusion}
& 12.9 & 1.4 & 5.1 & 1.5 & 10.7 & 1.4 \\
GaussianGraph~\cite{gaussiangraph}
& 20.7 & 12.1 & 10.9 & 6.3 & 17.2 & 8.6 \\
\midrule
\textbf{RelateGS (Ours)}
& \textbf{22.4} & \textbf{13.2}
& \textbf{21.1} & \textbf{6.4}
& \textbf{21.7} & \textbf{10.3} \\
\bottomrule
\end{tabular}
\label{tab:results_nr3d}
\end{table}

\begin{table}[!t]
\vspace{-0.8em}
\centering
\caption{
Semantic segmentation results on Replica~\cite{replica} and
ScanNet~\cite{scannet}.
We compare RelateGS with point-cloud and 3DGS-based scene representations.
}
\scriptsize
\setlength{\tabcolsep}{5pt}
\renewcommand{\arraystretch}{0.97}
\begin{tabular}{lcccc}
\toprule
\multirow{2}{*}{Method}
& \multicolumn{2}{c}{Replica}
& \multicolumn{2}{c}{ScanNet} \\
\cmidrule(lr){2-3}
\cmidrule(lr){4-5}
& mIoU$\uparrow$ & mAcc$\uparrow$
& mIoU$\uparrow$ & mAcc$\uparrow$ \\
\midrule
\multicolumn{5}{l}{\textit{Point-cloud-based}} \\
ConceptFusion~\cite{conceptfusion}
& 10.07 & 16.15 & 9.72 & 15.41 \\
ConceptGraph~\cite{conceptgraphs}
& 20.72 & 31.54 & 16.42 & 27.60 \\
HOV-SG~\cite{hierarchical}
& 23.16 & 29.85 & 22.43 & 43.81 \\
\midrule
\multicolumn{5}{l}{\textit{3DGS-based}} \\
LangSplat~\cite{langsplat}
& 4.72 & 9.12 & 3.28 & 8.95 \\
LEGaussian~\cite{LEGaussian}
& 4.80 & 11.59 & 3.51 & 10.04 \\
OpenGaussian~\cite{opengaussian}
& 26.39 & 44.28 & 24.73 & 41.54 \\
Occam's LGS~\cite{occamslgs}
& 40.21	& 59.97 & 30.80 & \textbf{49.31} \\
GaussianGraph~\cite{gaussiangraph}
& 31.18 & 49.14 & 31.09 & 48.91 \\
\midrule
\textbf{RelateGS (Ours)}
& \textbf{41.11} & \textbf{61.51}
& \textbf{32.70} & 49.04 \\
\bottomrule
\end{tabular}
\label{tab:semantic_segmentation}
\end{table}

\begin{table*}[!t]
\centering
\caption{
Open-vocabulary localization and segmentation results on the
\textbf{LERF-OVS} benchmark.
We report Acc@0.25, Acc@0.5, and mIoU (\%) for each scene.
Best and second-best results are highlighted in bold and underline,
respectively.
}
\scriptsize
\setlength{\tabcolsep}{3.0pt}
\renewcommand{\arraystretch}{0.97}
\begin{tabular}{lccc|ccc|ccc|ccc}
\toprule
\multirow{2}{*}{Method}
& \multicolumn{3}{c|}{ramen}
& \multicolumn{3}{c|}{teatime}
& \multicolumn{3}{c|}{waldo\_kitchen}
& \multicolumn{3}{c}{figurines} \\
\cmidrule(lr){2-4}
\cmidrule(lr){5-7}
\cmidrule(lr){8-10}
\cmidrule(lr){11-13}
& A@0.25$\uparrow$ & A@0.5$\uparrow$ & mIoU$\uparrow$
& A@0.25$\uparrow$ & A@0.5$\uparrow$ & mIoU$\uparrow$
& A@0.25$\uparrow$ & A@0.5$\uparrow$ & mIoU$\uparrow$
& A@0.25$\uparrow$ & A@0.5$\uparrow$ & mIoU$\uparrow$ \\
\midrule

LangSplat~\cite{langsplat}
& 11.09 & 7.25 & 8.94
& 21.73 & 19.05 & 14.42
& 10.92 & 10.08 & 10.79
& 9.00 & 7.31 & 14.84 \\

LEGaussian~\cite{LEGaussian}
& 31.54 & 26.76 & 15.79
& 30.49 & 27.12 & 19.27
& 25.76 & 18.18 & 11.78
& 29.41 & 23.21 & 17.99 \\

OpenGaussian~\cite{opengaussian}
& 40.85 & 21.13 & 25.41
& 79.66 & 71.19 & 59.00
& 40.15 & 35.45 & 25.15
& 78.57 & 71.43 & 57.50 \\

Occam's LGS~\cite{occamslgs}
& \underline{54.93} & \underline{45.07} & \underline{44.49}
& \textbf{89.83} & \textbf{81.36} & \textbf{71.03}
& \textbf{81.82} & \underline{59.09} & \underline{59.44}
& \underline{80.36} & \underline{60.71} & 59.85 \\

GaussianGraph~\cite{gaussiangraph}
& 43.66 & 28.63 & 29.58
& 79.96 & 76.27 & 62.70
& 51.42 & 36.36 & 35.85
& 81.24 & \textbf{75.61} & \textbf{62.00} \\

\textbf{RelateGS (Ours)}
& \textbf{57.75} & \textbf{47.89} & \textbf{46.11}
& \textbf{89.83} & \textbf{81.36} & \underline{70.20}
& \textbf{81.82} & \textbf{72.73} & \textbf{61.66}
& \textbf{83.93} & 58.93 & \underline{58.55} \\

\bottomrule
\end{tabular}
\label{tab:lerf_ovs}
\end{table*}

\noindent\textbf{Overall grounding performance.}
RelateGS consistently improves grounding over language-embedded Gaussian
representations and the structured GaussianGraph baseline.
On Sr3D+, RelateGS achieves 22.2 and 15.3 in overall Acc@0.1 and Acc@0.25,
respectively, compared with 18.2 and 7.4 for
GaussianGraph~\cite{gaussiangraph}.
This corresponds to improvements of 4.0 and 7.9 percentage points.
On Nr3D, RelateGS reaches 21.7/10.3 Acc@0.1/Acc@0.25, improving upon
GaussianGraph at 17.2/8.6.

Importantly, RelateGS uses the same downstream LLM-based graph matching
protocol as GaussianGraph.
The improvements therefore do not result from a new grounding decoder or a
stronger query-side reasoning module.
Instead, they demonstrate that refining the semantic and relational evidence
provided to the same matcher leads to more accurate target localization.

\noindent\textbf{Grounding under object ambiguity.}
The Easy and Hard splits further examine how the representation behaves when
the referred object must be distinguished from competing instances.
RelateGS shows improvements at the stricter Acc@0.25 criterion, including the challenging Hard subsets, indicating that refined relational
evidence contributes to more precise target localization.
The gains are not uniform across every metric, however, reflecting the
remaining difficulty of highly ambiguous referring expressions.
We provide additional per-scene results and failure cases in the
supplementary material.

\subsection{Evaluation of Semantic Evidence}
\label{sec:semantic_evaluation}

The grounding experiments evaluate the complete RelateGS representation.
We next examine the Gaussian-level semantic evidence independently of
relational reasoning.
These experiments are intended as supporting evaluations rather than
additional primary tasks: they test whether confidence-aware semantic
uplifting preserves discriminative language-aligned evidence before the
persistent object graph is used for grounding.

\noindent\textbf{Semantic segmentation.}
We evaluate semantic segmentation on Replica and ScanNet using mIoU and mAcc. RelateGS achieves strong semantic segmentation performance on both Replica
and ScanNet.
These results indicate that the Gaussian-level representation retains
discriminative semantic information while serving as the semantic basis of
the subsequent object-level representation.
The ScanNet results are particularly relevant because they evaluate the
semantic representation in the same type of cluttered indoor scenes used for
our grounding experiments.

\noindent\textbf{Open-vocabulary localization.}
We further evaluate language-to-region correspondence on LERF-OVS.
Table~\ref{tab:lerf_ovs} reports Acc@0.25, Acc@0.5, and mIoU for the four
benchmark scenes.
RelateGS remains competitive with specialized language-embedded Gaussian
representations across the four scenes.
It achieves the best results across all metrics on \textit{ramen} and strong
Acc@0.5 and mIoU performance on \textit{waldo\_kitchen}.
On \textit{teatime}, RelateGS matches the best localization accuracy while
remaining competitive in mIoU, whereas performance is more mixed on
\textit{figurines}.
Overall, the results show that the semantic representation used by RelateGS
maintains strong language-to-region correspondence while supporting the
object-centric relational representation required for downstream grounding.

\subsection{Ablation Studies}
\label{sec:ablation}

We conduct controlled ablations on Sr3D+ to isolate the two forms of
selective evidence refinement in RelateGS.
All variants use the same underlying Gaussian reconstruction and downstream
grounding protocol.
For variants involving relational re-querying, we additionally fix the
number of VLM queries so that improvements cannot be attributed simply to
additional inference.

\begin{table}[!t]
\centering
\caption{
Ablation of selective evidence refinement on Sr3D+.
CA denotes confidence-aware semantic uplifting and RQ denotes relational
re-querying.
Random and Graph indicate the strategy used to select re-query observations.
All RQ variants use the same additional VLM query budget.
}
\scriptsize
\setlength{\tabcolsep}{3.5pt}
\renewcommand{\arraystretch}{0.97}
\begin{tabular}{lccc|cc}
\toprule
Method
& CA & RQ & Selection
& A@0.1$\uparrow$ & A@0.25$\uparrow$ \\
\midrule
Initial Object Graph
& -- & -- & --
& 21.34 & 12.12 \\
+ Confidence-Aware Semantics
& \checkmark & -- & --
& 21.72 & 13.10 \\
+ Random Frame Re-query
& \checkmark & \checkmark & Random
& 19.78 & 11.62 \\
+ Graph-Guided Frame Re-query
& \checkmark & \checkmark & Graph
& \textbf{22.22} & \textbf{15.28} \\
\bottomrule
\end{tabular}
\label{tab:ablation_components}
\end{table}

\noindent\textbf{Effects of confidence-aware semantic uplifting.}
Adding confidence-aware observation selection to the initial object graph
improves Acc@0.1 from 21.34 to 21.72 and Acc@0.25 from 12.12 to 13.10.
The improvement therefore isolates the effect of filtering weak semantic
observations during Gaussian-level feature uplifting.
This result supports our premise that the quality of the multi-view evidence,
rather than merely its quantity, affects downstream grounding.

\noindent\textbf{Effects of additional relational re-querying.}
We next test whether additional VLM queries alone are sufficient to improve
grounding.
Random Frame Re-query uses the same additional query budget as our full
method but selects visible object pairs and observations without considering
deficiencies in the current graph.
Despite the additional observations, performance decreases to 19.78/11.62.
This result shows that uniformly adding relational observations is not
necessarily beneficial and may introduce evidence that is irrelevant or
ambiguous for target disambiguation.

\noindent\textbf{Effects of graph-guided relational refinement.}
Using the persistent graph to identify under-observed object pairs and
geometry to retrieve informative observations yields the best performance,
reaching 22.22/15.28 Acc@0.1/Acc@0.25.
Compared with Random Frame Re-query under the same VLM query budget, the
graph-guided strategy improves Acc@0.1 by 2.44 points and Acc@0.25 by
3.66 points.
These results directly support the central design of RelateGS: additional
evidence is most useful when its acquisition is conditioned on what the
current scene representation is missing.

\begin{figure*}[!t]
\vspace{-6pt}
\centering
\includegraphics[width=\linewidth]{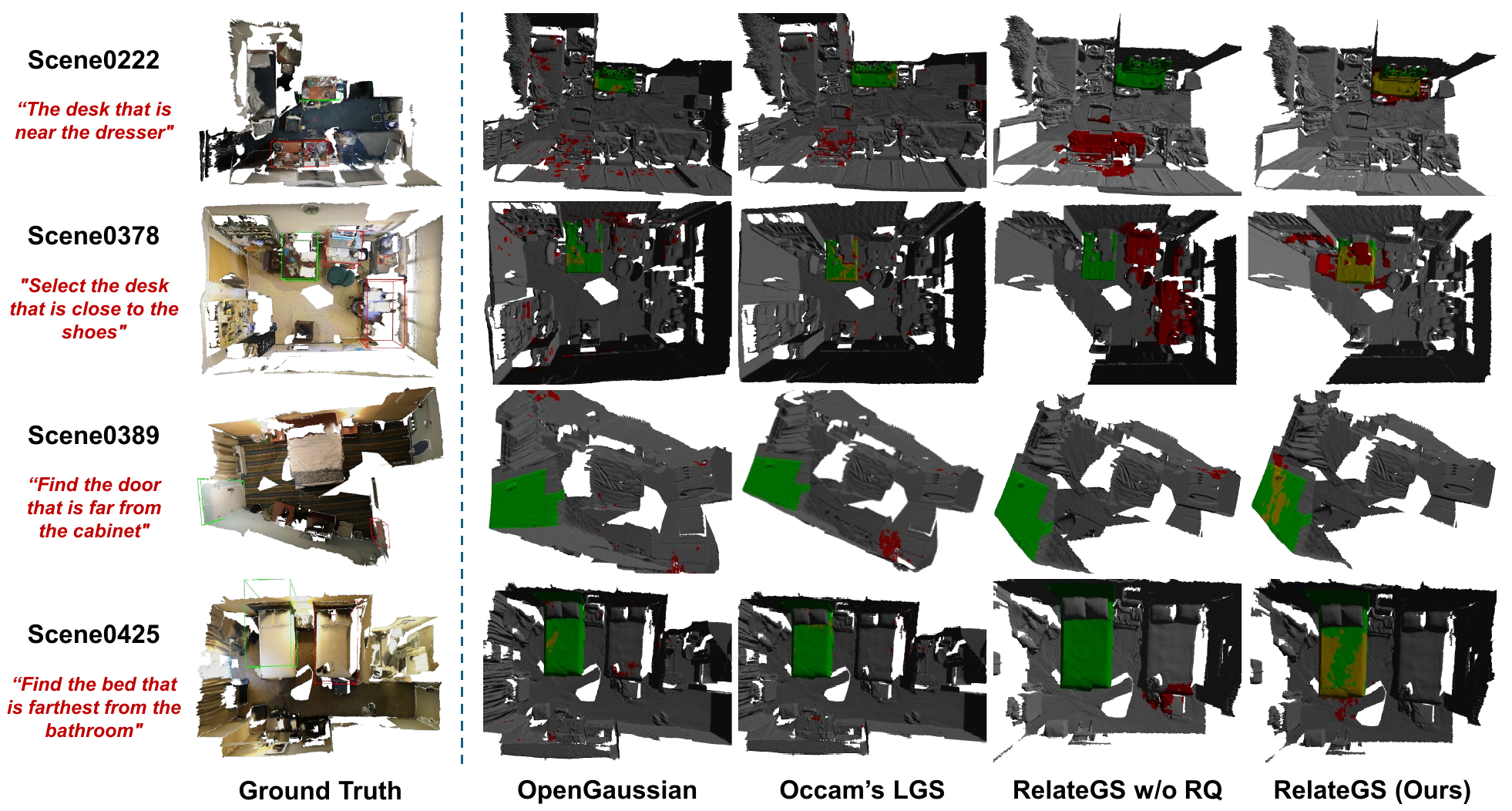}
\caption{
Qualitative comparison of relation-aware 3D visual grounding on ScanNet.
The ground-truth column shows the target object in green and distractor
objects in red.
For each method, green denotes the ground-truth target, yellow indicates a
correctly grounded prediction, and red indicates an incorrect prediction.
Compared with language-embedded Gaussian representations and RelateGS without
relational re-querying (w/o RQ), the full RelateGS more reliably resolves
targets from relational expressions by complementing under-observed
relational evidence.
}
\label{fig:main_experimental_vg_scannet}
\vspace{-6pt}
\end{figure*}

\begin{figure}[!t]
\vspace{-6pt}
\centering
\includegraphics[width=\linewidth]{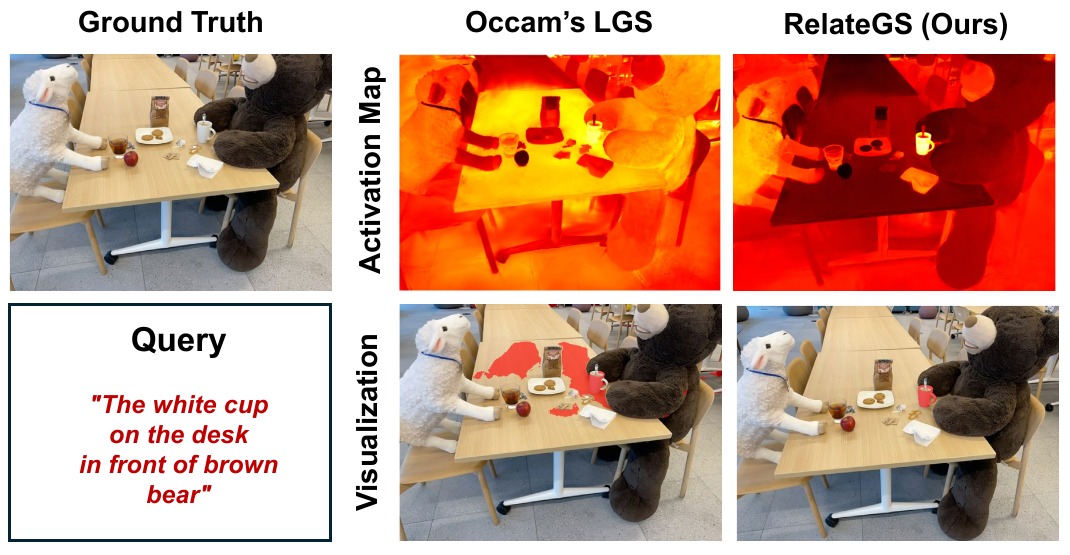}
\caption{
Qualitative comparison of open-vocabulary localization on LERF-OVS.
Given the same language query, we compare the activation maps and localized regions produced by Occam's LGS and RelateGS.
Confidence-aware semantic uplifting suppresses weak multi-view observations,
yielding more concentrated language responses around the referred object.
}
\label{fig:main_experimental_vg_lerf}
\vspace{-10pt}
\end{figure}

\subsection{Qualitative Results}
\label{sec:qualitative}

\noindent\textbf{Semantic evidence visualization.}
Figure~\ref{fig:main_experimental_vg_lerf} visualizes the language responses
of Occam's LGS and RelateGS for open-vocabulary localization.
While aggregating all multi-view observations can produce diffuse responses
over semantically irrelevant regions, confidence-aware uplifting suppresses
weak observations and produces a more concentrated response around the
referred object.
This result qualitatively supports the role of reliable Gaussian-level
semantic evidence in preserving discriminative language-to-region
correspondence.

\noindent\textbf{Relation-aware 3D grounding.}
Figure~\ref{fig:main_experimental_vg_scannet} compares 3D grounding results
for relational referring expressions on ScanNet.
These examples contain semantically similar distractors that cannot be
reliably resolved from category information alone.
While language-embedded Gaussian baselines often select incorrect instances, the persistent object representation provides explicit
inter-object relational evidence for target disambiguation.
Comparing RelateGS w/o RQ with the full model further shows the benefit of
graph-guided relational refinement: targeted re-querying complements
under-observed relations and enables the same downstream matcher to recover
the correct target in challenging cases.
\section{Conclusion}
\label{sec:conclusion}

We presented RelateGS, a structured language-aware Gaussian representation
for relation-aware 3D visual grounding.
Our key insight is that effective relational grounding requires not only
explicit object-level structure, but also reliable semantic and relational
evidence supporting that structure.
RelateGS addresses these complementary evidence failures by suppressing weak
multi-view observations during semantic feature uplifting and by using a
persistent object graph to identify and complement under-observed relations
through targeted re-querying of informative existing frames.
Experiments on Sr3D+ and Nr3D demonstrate the effectiveness of the resulting
representation for 3D visual grounding, while semantic segmentation and
open-vocabulary localization further validate the underlying language-aware
Gaussian representation.
Together, these results demonstrate the value of selectively refining
scene-side evidence for extending language-aware Gaussian representations
from semantic localization toward structured, relation-aware 3D scene
understanding.

{
    \small
    \bibliographystyle{ieeenat_fullname}
    \bibliography{main}

@String(CVPR= {IEEE Conf. Comput. Vis. Pattern Recog.})

@String(ICCV= {Int. Conf. Comput. Vis.})

@String(ECCV= {Eur. Conf. Comput. Vis.})

@String(BMVC= {Brit. Mach. Vis. Conf.})

@String(AAAI = {AAAI})

@String(CVPR  = {CVPR})

@String(ICCV  = {ICCV})

@String(ECCV  = {ECCV})

@String(BMVC  =	{BMVC})

@inproceedings{gaussiangraph,
    title = {{GaussianGraph}: 3D Gaussian-Based Scene Graph Generation for Open-World Scene Understanding},
    author = {Wang, Xihan and Yang, Dianyi and Gao, Yu and Yue, Yufeng and Yang, Yi and Fu, Mengyin},
    booktitle = {Proceedings of the IEEE/RSJ International Conference on Intelligent Robots and Systems (IROS)},
    pages = {4091--4098},
    year = {2025},
    doi = {10.1109/IROS60139.2025.11246515}
}

@inproceedings{hierarchical,
  author    = {Werby, Abdelrhman and Huang, Chenguang and B{\"u}chner, Martin and Valada, Abhinav and Burgard, Wolfram},
  title     = {Hierarchical Open-Vocabulary 3D Scene Graphs for Language-Grounded Robot Navigation},
  booktitle = {Proceedings of Robotics: Science and Systems},
  year      = {2024},
  address   = {Delft, Netherlands},
  month     = jul,
  doi       = {10.15607/RSS.2024.XX.077}
}

@inproceedings{conceptgraphs,
  title={Conceptgraphs: Open-vocabulary 3d scene graphs for perception and planning},
  author={Gu, Qiao and Kuwajerwala, Ali and Morin, Sacha and Jatavallabhula, Krishna Murthy and Sen, Bipasha and Agarwal, Aditya and Rivera, Corban and Paul, William and Ellis, Kirsty and Chellappa, Rama and others},
  booktitle={2024 IEEE International Conference on Robotics and Automation (ICRA)},
  pages={5021--5028},
  year={2024},
  organization={IEEE}
}

@inproceedings{sam,
  title={Segment anything},
  author={Kirillov, Alexander and Mintun, Eric and Ravi, Nikhila and Mao, Hanzi and Rolland, Chloe and Gustafson, Laura and Xiao, Tete and Whitehead, Spencer and Berg, Alexander C and Lo, Wan-Yen and others},
  booktitle={Proceedings of the IEEE/CVF international conference on computer vision},
  pages={4015--4026},
  year={2023}
}

@inproceedings{clip,
  title={Learning transferable visual models from natural language supervision},
  author={Radford, Alec and Kim, Jong Wook and Hallacy, Chris and Ramesh, Aditya and Goh, Gabriel and Agarwal, Sandhini and Sastry, Girish and Askell, Amanda and Mishkin, Pamela and Clark, Jack and others},
  booktitle={International conference on machine learning},
  pages={8748--8763},
  year={2021},
  organization={PmLR}
}

@article{3dgs,
  title={3d gaussian splatting for real-time radiance field rendering.},
  author={Kerbl, Bernhard and Kopanas, Georgios and Leimk{\"u}hler, Thomas and Drettakis, George},
  journal={ACM Trans. Graph.},
  volume={42},
  number={4},
  pages={139--1},
  year={2023}
}

@inproceedings{langsplat,
  title={Langsplat: 3d language gaussian splatting},
  author={Qin, Minghan and Li, Wanhua and Zhou, Jiawei and Wang, Haoqian and Pfister, Hanspeter},
  booktitle={Proceedings of the IEEE/CVF Conference on Computer Vision and Pattern Recognition},
  pages={20051--20060},
  year={2024}
}

@inproceedings{opengaussian,
  title={OpenGaussian: Towards Point-Level 3D Gaussian-Based Open Vocabulary Understanding},
  author={Wu, Yanmin and Meng, Jiarui and Li, Haijie and Wu, Chenming and Shi, Yahao and Cheng, Xinhua and Zhao, Chen and Feng, Haocheng and Ding, Errui and Wang, Jingdong and Zhang, Jian},
  booktitle={Advances in Neural Information Processing Systems},
  volume={37},
  pages={19114--19138},
  year={2024}
}

@inproceedings{LEGaussian,
  title={Language embedded 3d gaussians for open-vocabulary scene understanding},
  author={Shi, Jin-Chuan and Wang, Miao and Duan, Hao-Bin and Guan, Shao-Hua},
  booktitle={Proceedings of the IEEE/CVF Conference on Computer Vision and Pattern Recognition},
  pages={5333--5343},
  year={2024}
}

@inproceedings{bbq,
  author    = {Linok, Sergey and Zemskova, Tatiana and
               Ladanova, Svetlana and Titkov, Roman and
               Yudin, Dmitry and Monastyrny, Maxim and
               Valenkov, Aleksei},
  title     = {Beyond Bare Queries: Open-Vocabulary Object
               Grounding with 3D Scene Graph},
  booktitle = {Proceedings of the 2025 IEEE International
               Conference on Robotics and Automation (ICRA)},
  pages     = {13582--13589},
  year      = {2025},
  doi       = {10.1109/ICRA55743.2025.11128059}
}

@inproceedings{openfusion,
  title={Open-fusion: Real-time open-vocabulary 3d mapping and queryable scene representation},
  author={Yamazaki, Kashu and Hanyu, Taisei and Vo, Khoa and Pham, Thang and Tran, Minh and Doretto, Gianfranco and Nguyen, Anh and Le, Ngan},
  booktitle={2024 IEEE International Conference on Robotics and Automation (ICRA)},
  pages={9411--9417},
  year={2024},
  organization={IEEE}
}

@inproceedings{referit3d,
  title={Referit3d: Neural listeners for fine-grained 3d object identification in real-world scenes},
  author={Achlioptas, Panos and Abdelreheem, Ahmed and Xia, Fei and Elhoseiny, Mohamed and Guibas, Leonidas},
  booktitle={Computer Vision--ECCV 2020: 16th European Conference, Glasgow, UK, August 23--28, 2020, Proceedings, Part I 16},
  pages={422--440},
  year={2020},
  organization={Springer}
}

@article{replica,
  title={The replica dataset: A digital replica of indoor spaces},
  author={Straub, Julian and Whelan, Thomas and Ma, Lingni and Chen, Yufan and Wijmans, Erik and Green, Simon and Engel, Jakob J and Mur-Artal, Raul and Ren, Carl and Verma, Shobhit and others},
  journal={arXiv preprint arXiv:1906.05797},
  year={2019}
}

@inproceedings{scannet,
  title={Scannet: Richly-annotated 3d reconstructions of indoor scenes},
  author={Dai, Angela and Chang, Angel X and Savva, Manolis and Halber, Maciej and Funkhouser, Thomas and Nie{\ss}ner, Matthias},
  booktitle={Proceedings of the IEEE conference on computer vision and pattern recognition},
  pages={5828--5839},
  year={2017}
}

@article{conceptfusion,
  title={Conceptfusion: Open-set multimodal 3d mapping},
  author={Jatavallabhula, Krishna Murthy and Kuwajerwala, Alihusein and Gu, Qiao and Omama, Mohd and Chen, Tao and Maalouf, Alaa and Li, Shuang and Iyer, Ganesh and Saryazdi, Soroush and Keetha, Nikhil and others},
  journal={arXiv preprint arXiv:2302.07241},
  year={2023}
}

@inproceedings{reasongrounder,
  author    = {Liu, Zhenyang and Wang, Yikai and Zheng, Sixiao
               and Pan, Tongying and Liang, Longfei
               and Fu, Yanwei and Xue, Xiangyang},
  title     = {ReasonGrounder: LVLM-Guided Hierarchical Feature
               Splatting for Open-Vocabulary 3D Visual Grounding
               and Reasoning},
  booktitle = {Proceedings of the IEEE/CVF Conference on Computer
               Vision and Pattern Recognition (CVPR)},
  pages     = {3718--3727},
  year      = {2025},
  doi       = {10.1109/CVPR52734.2025.00352}
}

@inproceedings{core3dvg,
  author    = {Yang, Li and Yuan, Chunfeng and Zhang, Ziqi
               and Qi, Zhongang and Xu, Yan and Liu, Wei
               and Shan, Ying and Li, Bing and Yang, Weiping
               and Li, Peng and Wang, Yan and Hu, Weiming},
  title     = {Exploiting Contextual Objects and Relations
               for 3D Visual Grounding},
  booktitle = {Advances in Neural Information Processing Systems},
  volume    = {36},
  year      = {2023},
  url       = {https://papers.nips.cc/paper_files/paper/2023/hash/9b91ee0da3bcd61905fcd89e770168fc-Abstract-Conference.html}
}

@misc{viewsrd,
      title={ViewSRD: 3D Visual Grounding via Structured Multi-View Decomposition}, 
      author={Ronggang Huang and Haoxin Yang and Yan Cai and Xuemiao Xu and Huaidong Zhang and Shengfeng He},
      year={2025},
      eprint={2507.11261},
      archivePrefix={arXiv},
      primaryClass={cs.CV},
      url={https://arxiv.org/abs/2507.11261}, 
}

@inproceedings{3dssg,
  title={Learning 3d semantic scene graphs from 3d indoor reconstructions},
  author={Wald, Johanna and Dhamo, Helisa and Navab, Nassir and Tombari, Federico},
  booktitle={Proceedings of the IEEE/CVF Conference on Computer Vision and Pattern Recognition},
  pages={3961--3970},
  year={2020}
}

@inproceedings{scenegraphfusion,
  title={Scenegraphfusion: Incremental 3d scene graph prediction from rgb-d sequences},
  author={Wu, Shun-Cheng and Wald, Johanna and Tateno, Keisuke and Navab, Nassir and Tombari, Federico},
  booktitle={Proceedings of the IEEE/CVF Conference on Computer Vision and Pattern Recognition},
  pages={7515--7525},
  year={2021}
}

@inproceedings{vlsat,
  title={Vl-sat: Visual-linguistic semantics assisted training for 3d semantic scene graph prediction in point cloud},
  author={Wang, Ziqin and Cheng, Bowen and Zhao, Lichen and Xu, Dong and Tang, Yang and Sheng, Lu},
  booktitle={Proceedings of the IEEE/CVF conference on computer vision and pattern recognition},
  pages={21560--21569},
  year={2023}
}

@inproceedings{sgaligner,
  title={Sgaligner: 3d scene alignment with scene graphs},
  author={Sarkar, Sayan Deb and Miksik, Ondrej and Pollefeys, Marc and Barath, Daniel and Armeni, Iro},
  booktitle={Proceedings of the IEEE/CVF International Conference on Computer Vision},
  pages={21927--21937},
  year={2023}
}

@inproceedings{occamslgs,
  author    = {Cheng, Jiahuan and Zaech, Jan-Nico
               and Van Gool, Luc and Paudel, Danda Pani},
  title     = {Occam's LGS: An Efficient Approach for
               Language Gaussian Splatting},
  booktitle = {Proceedings of the 36th British Machine
               Vision Conference (BMVC)},
  year      = {2025},
  publisher = {BMVA},
  url       = {https://bmva-archive.org.uk/bmvc/2025/assets/papers/Paper_694/paper.pdf}
}

@inproceedings{llavaspacesgg,
    title = {{LLaVA-SpaceSGG}: Visual Instruct Tuning for Open-Vocabulary Scene Graph Generation with Enhanced Spatial Relations},
    author = {Xu, Mingjie and Wu, Mengyang and Zhao, Yuzhi and Li, Jason Chun Lok and Ou, Weifeng},
    booktitle = {Proceedings of the Winter Conference on Applications of Computer Vision (WACV)},
    month = {February},
    year = {2025},
    pages = {6362--6372},
    doi = {10.1109/WACV61041.2025.00620}
}

@inproceedings{ludvig,
            title={LUDVIG: Learning-free Uplifting of 2d Visual features to Gaussian Splatting scenes},
            author={Marrie, Juliette and Menegaux, Romain and Arbel, Michael and Larlus, Diane and Mairal, Julien},
            booktitle={Proceedings of the IEEE/CVF International Conference on Computer Vision (ICCV)},
            year={2025}
        }

@inproceedings{eg3dvg,
  title={EG-3DVG: Expression and Geometry Aware Grounding Decoder for 3D Visual Grounding},
  author={Park, GwangWook and Lee, Hyo-Jun and Baek, Jong-Hyeon and Kim, Hanul and Koh, Yeong Jun},
  booktitle={Proceedings of the IEEE/CVF Conference on Computer Vision and Pattern Recognition},
  pages={2625--2634},
  year={2026}
}

@inproceedings{chainsemantics,
  title={Chain of semantics programming in 3d gaussian splatting representation for 3d vision grounding},
  author={Shi, Jiaxin and Xiang, Mingyue and Sun, Hao and Huang, Yixuan and Weng, Zhi},
  booktitle={2025 IEEE/CVF Conference on Computer Vision and Pattern Recognition (CVPR)},
  pages={24560--24569},
  year={2025},
  organization={IEEE}
}

@InProceedings{concretenet,
      author = {Unal, Ozan and Sakaridis, Christos and Saha, Suman and Van Gool, Luc},
      title = {Four Ways to Improve Verbo-visual Fusion for Dense 3D Visual Grounding},
      booktitle = {European Conference on Computer Vision (ECCV)},
      month = {October},
      year = {2024}
      }

@inproceedings{gags,
  title     = {GAGS: Granularity-Aware Feature Distillation for Language Gaussian Splatting},
  author    = {Peng, Yuning and Wang, Haiping and Liu, Yuan and Wen, Chenglu and Dong, Zhen and Yang, Bisheng},
  booktitle = {Proceedings of the AAAI Conference on Artificial Intelligence},
  volume    = {40},
  number    = {10},
  pages     = {8376--8384},
  year      = {2026},
  doi       = {10.1609/aaai.v40i10.37787}
}
}

\end{document}